# Mitigating Biases in Student Performance Prediction via Attention-Based Personalized Federated Learning

Yun-Wei Chu[†], Seyyedali Hosseinalipour[†], Elizabeth Tenorio[†], Laura Cruz[†],
Kerrie Douglas[†], Andrew Lan[◦], Christopher Brinton[†]
[†]Purdue University, [◦]University of Massachusetts Amherst
{chu198,hosseina,etenori,lcruzcas,douglask,cgb}@purdue.edu,andrewlan@cs.umass.edu

## ABSTRACT
Traditional learning-based approaches to student modeling generalize poorly to underrepresented student groups due to biases in data availability. In this paper, we propose a methodology for predicting student performance from their online learning activities that optimizes inference accuracy over different demographic groups such as race and gender. Building upon recent foundations in federated learning, in our approach, personalized models for individual student subgroups are derived from a global model aggregated across all student models via meta-gradient updates that account for subgroup heterogeneity. To learn better representations of student activity, we augment our approach with a self-supervised behavioral pretraining methodology that leverages multiple modalities of student behavior (e.g., visits to lecture videos and participation on forums), and include a neural network attention mechanism in the model aggregation stage. Through experiments on three real-world datasets from online courses, we demonstrate that our approach obtains substantial improvements over existing student modeling baselines in predicting student learning outcomes for all subgroups. Visual analysis of the resulting student embeddings confirm that our personalization methodology indeed identifies different activity patterns within different subgroups, consistent with its stronger inference ability compared with the baselines.

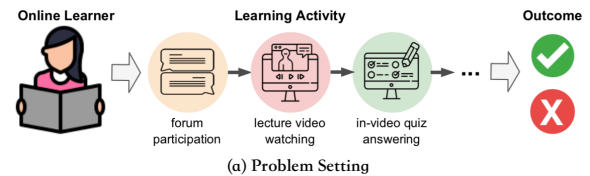

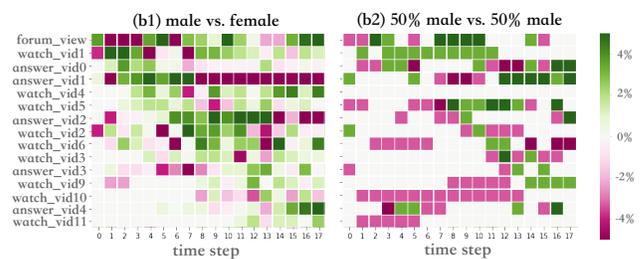

Figure 1: (a) Our problem setting of extracting online student learning activity to predict learning outcomes. (b) Heatmaps of differences in learning activities exhibited (b1) between different student subgroups, male vs. female, and (b2) within the male subgroup for the QDS dataset (see Section 2).

## 1 INTRODUCTION
Online learning [44] has proven to be an important component of today's educational processes, highlighted by its significant uptick in usage during the COVID-19 pandemic [1]. Due to the remote nature of online learning, it is harder for instructors to pay attention to each individual student to provide feedback and to deliver personalized learning experiences. This has motivated the investigation of artificial intelligence (AI)-based approaches to delivering personalized instruction and feedback based on measured student progress in online learning activities [4, 26, 33].

*Student modeling* [37] is an important research area since it provides information on key individual student factors that drive personalization systems. There exist a wide range of student models, from those that analyze student *knowledge*, such as item response theory [36] and models for knowledge tracing [14], to those that analyze student *behavior* within computer-based learning platforms to detect psychological states [6] and discover learning tendencies [11, 41, 42]. Since these student models are fitted from actual student data collected from real-world learning platforms, they are inherently prone to any biases that exist in the available data [17]. The research topic of de-biasing data-driven student models has gained significant recent traction; there exist studies that investigate existing algorithmic biases in educational applications [18, 23] and explorations on how to impose constraints during model training to promote fairness across different student groups [43].

One common theme of these works is that they study the bias/fairness aspects of a single *global* student model that is trained on data collected from *all* students [31]. This setup is typically effective in AI applications since more data generally leads to improved model fit. However, this setup ignores the fact that *underrepresented minorities* may not be well-captured by a population-level model, resulting in unfair predictions that may cause catastrophic errors [3, 10, 25]. On the contrary, training separate *local* models for each student subgroup may not be effective since small subgroups do not Yun-Wei Chu[†], Seyyedali Hosseinalipour[†], Elizabeth Tenorio[†], Laura Cruz[†], Kerrie Douglas[†], Andrew Lan[◦], Christopher Brinton[†] have enough data for us to train an accurate model. In this work, we aim to develop a *personalized student modeling methodology* that addresses the data availability challenges across subgroups.

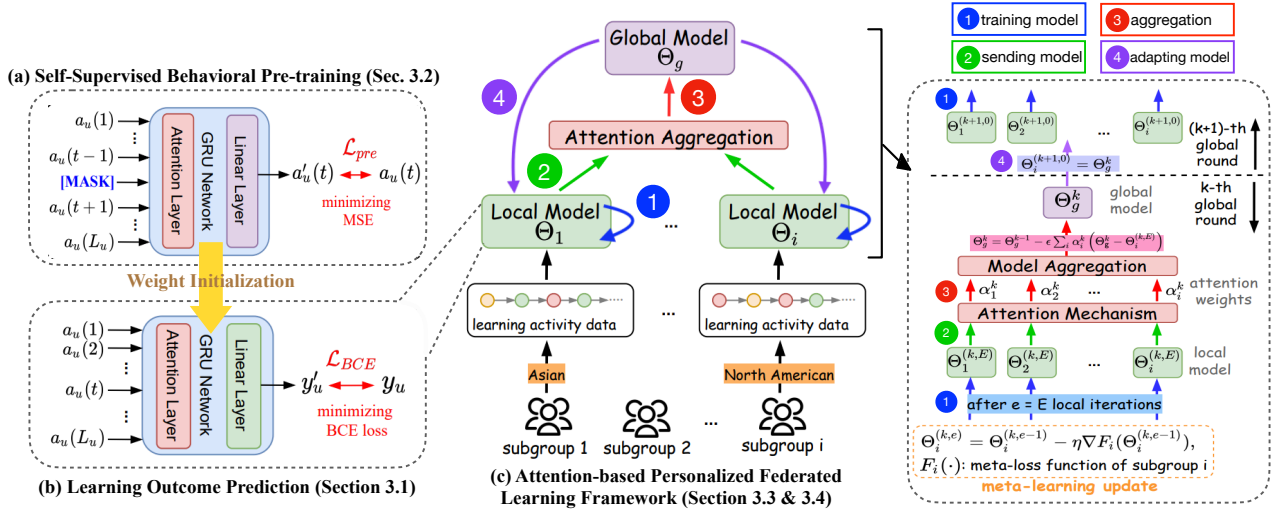

Figure 2: Overview of the personalized prediction methodology we develop in this paper for customizing prediction models for each demographic student subgroup. After encoding the activity data (Sec. 2), the model first learns a surrogate objective by predicting the activity type in a self-supervised manner (Sec. 3.2). Based on this pre-training knowledge, a time-series neural network models students' learning trajectories and predicts their learning performance (Sec. 3.1). Finally, a personalized federated learning framework (Sec. 3.3 and 3.4) trains local student models that are personalized to each student subgroup by adapting from a shared global model to both leverage commonality and maintain heterogeneity between subgroups.

## 1.1 Federated Learning and Student Modeling

The recently proposed federated deep knowledge tracing (FDKT) [40] method is a first step towards coordinating between global and local student models, based on federated learning (FL) [24]. While FL was traditionally envisioned for distributed systems, its main idea is more general: enabling collaboration among models trained on local datasets to achieve better performance collectively. Research in *Global FL* has proposed various aggregation strategies to optimize the discovered global model, e.g., FedAvg [28] and FedAtt [20]. However, statistical heterogeneity across local data distributions presents fundamental challenges to Global FL [19], which has motivated approaches for *Personalized FL* to adapt local models [16, 21].

FDKT considers statistical heterogeneity arising from student training data being split across different schools, where the schools are not willing to share data due to privacy concerns. Thus, local models are trained for each school, and updating/averaging between the global and local models is coordinated without moving the actual data. While this setup is representative of some real-world educational settings, it ignores the bigger picture of heterogeneous student subgroups according to race, gender, and other demographic variables. The aggregation scheme in FDKT weights the local models according to "*data quality*," measured through fitting local psychometric models such as classical test theory and item response theory. Down weighting a student subgroup is not acceptable in our setting since the data from each subgroup contains important information about how these students learn.

FDKT's personalization strategy relies on heuristic linear interpolation to adapt the global model to each local model, resembling existing techniques in domain adaptation [5, 15]. The interpolation coefficient is calculated through a rule set of priori, and thus may not strike the right balance between these models. Our personalization approach instead builds upon *federated meta learning* [16], a recent innovation in FL which refines the global model to encapsulate local adaptation through meta-functions. As we will see, managing local models for each student subgroup under this framework provides significant enhancements in prediction quality across subgroups.

## 1.2 Activity-Based Performance Prediction

In this work, we focus on the downstream application of *student outcome prediction* in online courses. Our approach will build upon and formalize two key observations from past educational data mining research. First is that patterns in student learning activity (e.g., video-watching behavior, discussion forum interactions) contain signals of student performance in online courses [7, 9, 13]. Since each student tends to generate a substantial number of activity measurements, incorporating them may alleviate sparsity issues faced by prediction models trained on subgroups with fewer students. Figure 1(a) visualizes our goal of predicting a student's final course grade from various online learning activities captured.

The second observation is that distinctive patterns of learning activities can be identified both within and across demographic subgroups [2, 12, 30]. This was originally discovered among in-person learning behaviors (e.g., participation and learning style), and more recently in digital environments [27]. We thus aim to capture how sequences of learning activities (e.g., viewing a forum, then watching a specific video, then answering a specific quiz question) differ based on student subgroups in our construction of personalized FL models. However, online learning activities contain more noise than those capture in-person (e.g., a student accessing a video accidentally), making it difficult to identify these patterns through standard data mining techniques [7]. For example, considering one of the datasets in this paper, Figure 1(b) gives heatmaps of differences observed in when students are engaged in particular learning activities. Two cases are considered: (b1) between learners across a demographic group (all males vs. all females), and (b2) between learners within a subgroup (50% of the males vs. the other 50%, chosen randomly), with each value indicating the difference in the fraction of students engaged in the activity at that point in their learning trajectory. (b1) exhibits a more varied set of trajectories than (b2), consistent with the observation that different subgroups have different learning behaviors, but (b2) still has noticeable differentials even though it is a comparison among a cross-section of the same subgroup. This is further motivation for our meta-learning approach to subgroup personalization, where any machine-identified commonalities in the data across the subgroups are captured in a global model that is further refinable based on local subgroup information. As we will see in Sec. 4.5, visualizations of the student embeddings learned by our methodology validate that distinctive activity patterns can be extracted between subgroups.

### 1.3 Summary of Contributions

The methodology that we develop in this paper is depicted at a high level in Figure 2. Specifically, we make three major contributions:

- We develop a personalized federated learning methodology (Sec. 3.3 & 3.4) where local models associated with different student subgroups are adapted from the global model. This adaptation is in the form of meta-gradient updates taken on data that is localized to the specific student subgroup, instead of using data quality heuristics as in prior work.
- We develop a behavior-based local modeling architecture for predicting student performance from multi-modal learning activity logs (Sec. 2 & 3.1). We further show that a self-supervised pre-training process leads to better activity representations and improvements in prediction performance.
- Through extensive experiments on three real-world online courses datasets (Sec. 4), we demonstrate that our methodology leads to a set of student models that significantly outperform existing approaches to performance prediction. We find this advantage appears at both the global and local levels, resulting in a prediction model that can be quickly adapted to different student subgroups.

## 2 DATA ENCODING METHODOLOGY

In this section, we formalize our methodology for encoding student activity and demographic data, which will be employed in our student modeling techniques in Section 3. We collect our datasets from Purdue University's offerings of online courses on the edX platform (www.edx.org). We study three graduate-level online certificate courses: Fiber Optic Communications (FOC), Quantum Detectors and Sensors (QDS), and Essentials of MOSFETs (MOSFETs). Each course consists of a series of lecture videos, some of which have an end-of-video quiz that assesses student learning progress. Moreover, each course provides a discussion forum page for students to discuss with each other. The platform also records each student's final grade, which is a pass or fail label based on the grading policy. Summary statistics of the three courses are given in Table 1, emphasizing a broad range of activity levels for our evaluation.

|  | FOC | QDS | MOSFETs |
|---|---|---|---|
| # of students | 1,265 | 2,304 | 886 |
| # of lecture videos | 43 | 31 | 26 |
| # end-of-video quizzes | 25 | 23 | 11 |
| # of discussion threads | 20 | 35 | 17 |
| Avg. reply per thread | 1.95 | 0.48 | 1.44 |
| # of activities | 63,789 | 95,487 | 31,382 |
| video activity vs. forum activity | 67%, 33% | 70%, 30% | 59%, 41% |
| final pass rate | 29.3% | 24.8% | 31.1% |

Table 1: Basic information on datasets collected from edX.

### 2.1 Video-watching Activity

Each time a student $u$ accesses a lecture video $v$, their activity is recorded with the following information: student ID, video ID, and UNIX timestamp. The in-video quizzes for each of the courses consist of a single multiple-choice question or a True/False question, appearing at the end of the video. When student $u$ submits an answer to the in-video question for video $v$, the platform records the student's answer, the points rewarded $o_{u,v}$, and the maximum possible points for this question $o_v^{\max}$. A common metric for student performance on quiz questions is whether they were correct on their first attempt or not [7, 9, 13]. Therefore, we consider the score that student $u$ obtained at the end of lecture video $v$ as $s_{u,v} = 1$ when $o_{u,v} = o_v^{\max}$; otherwise, $s_{u,v} = 0$. The video-watching activities can then be divided into four types according to the quiz outcome:

(1) watch_noquiz: The student watched the lecture video, but the video did not have an end-of-video quiz.
(2) watch_correct: The student watched the lecture video and answered the end-of-video quiz correctly, i.e., $s_{u,v} = 1$.
(3) watch_incorrect: The student watched the lecture video and answered the quiz incorrectly, i.e., $s_{u,v} = 0$.
(4) watch_noanswer: The student watched the lecture video but did not submit any answer to the end-of-video quiz.

### 2.2 Forum-participation Activity

Discussion forums are considered an extension of traditional learning that promotes dialogue, reflection, and knowledge construction [8]. They are particularly important in virtual learning environments, motivating us to consider them for student knowledge modeling. The learning platform records forum-participation activities when student $u$ visits the discussion forum page. Similar to video-watching, the learning platform records the student ID and UNIX timestamp, and also the text of the post/reply for each activity. We define three types of forum-participation activities:

(1) forum_post: The student either started a new thread or made a post in a thread.
(2) forum_reply: The student replied to another student's post.
(3) forum_view: The student visited a thread without posting or replying. Most of the forum-participation activities belong to the forum_view type. The platform cannot identify which thread a student views; nevertheless, this activity suggests that a student is seeking information through the forums.

### 2.3 Activity Encoding

We use one-hot encodings to represent the different activity types. We encode the video ID as $\mathbf{v}^{\text{id}} \in \{0, 1\}^n$, where $n$ denotes the number of videos, which are 43, 31, and 26 for the FOC, QDS, and Yun-Wei Chu[†], Seyyedali Hosseinalipour[†], Elizabeth Tenorio[†], Laura Cruz[†], Kerrie Douglas[†], Andrew Lan[°], Christopher Brinton[†]

| dataset | FOC | QDS | MOSFETs | FOC | QDS | MOSFETs |
|---|---|---|---|---|---|---|
| overall statistics | 1,265 | 2,304 | 886 | 29.3% | 24.8% | 28.5% |
| Gender (G) | | | | | | |
| Male ($\Omega^{G,M}$) | 42.7% | 42.9% | 43.2% | 27.7% | 31.8% | 33.6% |
| Female ($\Omega^{G,F}$) | 8.3% | 11.3% | 23.7% | 23.0% | 32.5% | 30.9% |
| Unspecified ($\Omega^{\neg G}$) | 49.0% | 45.8% | 33.1% | 27.2% | 31.8% | 30.1% |
| Continent (C) | | | | | | |
| Asian ($\Omega^{C,AS}$) | 35.0% | 42.8% | 41.4% | 31.4% | 34.5% | 36.7% |
| African ($\Omega^{C,AF}$) | 10.1% | 6.9% | 7.2% | 8.5% | 5.2% | 10.6% |
| European ($\Omega^{C,EU}$) | 14.4% | 13.3% | 14.6% | 11.2% | 9.9% | 17.7% |
| North American ($\Omega^{C,NA}$) | 24.5% | 21.9% | 21.9% | 19.8% | 18.3% | 28.1% |
| South American ($\Omega^{C,SA}$) | 5.5% | 4.2% | 3.9% | 4.4% | 2.8% | 10.0% |
| Unspecified ($\Omega^{\neg C}$) | 10.5% | 10.9% | 11.0% | 4.6% | 6.7% | 11.4% |
| Year of Birth (Y) | | | | | | |
| Y ≤ 1980 ($\Omega^{Y,\sim 80}$) | 12.1% | 9.0% | 14.0% | 18.7% | 29.5% | 27.9% |
| 1980 < Y ≤ 1990 ($\Omega^{Y,80\sim 90}$) | 17.9% | 16.2% | 15.2% | 26.5% | 28.2% | 34.8% |
| Y > 1990 ($\Omega^{Y,90\sim}$) | 22.8% | 29.8% | 31.5% | 32.7% | 23.7% | 30.9% |
| Unspecified ($\Omega^{\neg Y}$) | 42.7% | 45.0% | 39.3% | 28.8% | 22.1% | 28.6% |

**Table 2: (Left) Population distribution and (Right) pass rate distribution for each student subgroup in each course.**

MOSFET datasets, respectively. Video-watching activity is encoded as $\mathbf{a}^v \in \{0,1\}^4$ according to the four different activity types. Similarly, forum-participation activity is encoded as $\mathbf{a}^f \in \{0,1\}^3$. For each student $u$, their activity record at time step $t$ is defined as:

$$\mathbf{a}_u(t) = \left(\mathbf{v}^{id};\ \mathbf{a}^v;\ \mathbf{a}^f\right), \quad (1)$$

where (;) denotes vector concatenation. We concatenate video-watching and forum-participation activity in this way to ensure a consistent activity dimension. Note that these two activities do not happen simultaneously; whenever a student watches a video, the forum-participation part of this encoding, i.e., $\mathbf{a}^f = \mathbf{0}$, while $\mathbf{v}^{id}$ and $\mathbf{a}^v$ capture the activity. On the other hand, $\mathbf{v}^{id}, \mathbf{a}^v = \mathbf{0}$ when the student engages in a forum-participation activity. The time series $\{\mathbf{a}_u(t)\}$ of activities for each student are ordered by UNIX timestamp, i.e., in Figure 1(b), $t = 0, ..., 17$ are the first 18 activities.

### 2.4 Student Subgroups

edX solicits a student's demographic information when they register, which they can fill out voluntarily. We consider three demographic variables in our analysis: (i) gender[1], (ii) country, and (iii) year of birth. We group countries into continents[2] and year of birth into decades to have a reasonable sample size in each subgroup.

The distribution of students' demographic variables is summarized in Table 2. For each course, we group students into two subsets, $\Omega^I$ and $\Omega^{\neg I}$, based on each demographic variable $I \in \{G, C, Y\}$, where $G, C$, and $Y$ represent gender, continent, and year of birth, respectively. $\Omega^I$ is the set of students that voluntarily provided information on variable $I$ while $\Omega^{\neg I}$ denotes the set of students that chose not to provide this information. For the set $\Omega^I$, we further define $\Omega^{I,x}$ as the set of students belonging to subgroup $x \in \mathcal{X}$, where $\mathcal{X}$ is the set of groups in variable $I$. For example, $\mathcal{X} = \{M, F\}$ represents male and female subgroups for gender information ($I = G$) when it is provided. The pass rate distributions for each student subgroup are also summarized in Table 2.

---
[1] edX only provided a binary choice of male and female for selection.
[2] We group students into five continents according to their country information since no students claimed to be from Antarctica and countries in Oceania.

## 3 PERSONALIZED PREDICTION MODELING

We now present our prediction methodology based on the student activity encodings. Referring back to Figure 2, our base prediction model architecture will leverage recurrent neural networks (RNNs) given the sequential nature of the input data (Section 3.1). To augment these RNNs with prior knowledge on how students behave, we develop a self-supervised pre-training process (Section 3.2). Our personalized FL framework is depicted in Figure 2(c), consisting of a local-global architecture which customizes prediction models for each student subgroup through meta-learning (Section 3.3). After a few local updates, the global model is re-computed through an attention-based aggregation scheme (Section 3.4).

### 3.1 Prediction Model Structure

We leverage attention-based Gated Recurrent Units (GRU), known for their ability to capture dependencies over long time periods [13]. For each student $u$, our prediction model aims to predict a binary classification label $y_u \in \{0, 1\}$, representing whether the student passes the course or not, based on their time-dependent activity sequence $\mathbf{a}_u(t)$, where $t \in \{1, ..., L_u\}$ is the discrete time index and $L_u$ is the length of the time series sequence for user $u$.

The model takes encoded activities as an input, generates learned representations for each user, and outputs representations for predicting learning performance. The hidden state of the GRU model is formulated as:

$$\mathbf{h}_u(t) = \text{GRU}\left(\mathbf{a}_u(t),\ \mathbf{h}_u(t-1)\right). \quad (2)$$

Encoding a long time sequence into a single final state ($\mathbf{h}_u(L_u)$) might be unrealistic. To overcome this, [38] proposed a self-attention mechanism that weights the $\mathbf{h}_u(t)$ over time. Specifically, instead of forcing the network to encode all the information into the final state, an attention module takes all the $\mathbf{h}_u(t)$ as inputs and generates the learned representation $\widetilde{\mathbf{h}}_u$ as output. Adopting this idea for our setting, we define an attention module as:

$$\widetilde{\mathbf{h}}_u = \sum_t \alpha(t) \mathbf{h}_u(t), \quad (3)$$

where the weight $\alpha(t) = \frac{\exp(e(t))}{\sum_t \exp(e(t))}$, $e(t) = p(t)^\top \tanh\left(\mathbf{W}_\alpha \mathbf{h}_u(t)\right)$, and $\mathbf{W}_\alpha$ is a learned parameter. A linear layer then transforms the learned representation, and the predicted probability of pass/fail labels is defined as:

$$\mathbf{y}'_u = \text{softmax}\left(\mathbf{W}_l \cdot \widetilde{\mathbf{h}}_u + \mathbf{b}_l\right), \quad (4)$$

where $\mathbf{W}_l \in \mathbb{R}^{k \times 2}$ is the weight matrix for linearly transforming $\widetilde{\mathbf{h}}_u$, $k = 48$ is hidden dimension, and $\mathbf{b}_l \in \mathbb{R}^2$ is the bias vector. The parameters of our attention-based GRU model are initialized through the pre-training process discussed next. Finally, the loss function for evaluating the prediction quality is taken as binary cross entropy (BCE) loss $\mathcal{L}_{BCE}$:

$$\mathcal{L}_{BCE} = -\sum_{u \in \Omega} \mathbf{y}_u \log(\mathbf{y}'_u) + (\mathbf{1} - \mathbf{y}_u) \log(\mathbf{1} - \mathbf{y}'_u), \quad (5)$$

where $\mathbf{1}$ is an all-one vector, $\mathbf{y}'_u$ is the prediction of the model, and $\mathbf{y}_u$ is the one-hot encoding vector of the binary label $y_u$.

## 3.2 Self-Supervised Behavioral Pre-training

As discussed in Section 1, a key challenge we face is that the number of students in an underrepresented subgroup can be small. With a large amount of activity information per student, our methodology will benefit from a pre-training process that learns a representation of how students behave. Motivated by this, we design a pre-training method for our framework based on the concept of Continuous Bag Of Words (CBOW) [29], which has achieved notable success for NLP applications. In our setting, the representations of surrounding activities are combined to predict the current activity.

The pre-training architecture is shown in Figure 2(a). We extract each activity $\mathbf{a}_u(t)$ as a target and train the model by taking the rest of the activities as an input. Note that we inject a zero vector at the position where the activity is taken as the prediction target. The model architecture employed (GRU with attention) is the same as in the previous section. The model generates the final hidden state $\mathbf{h}_{\text{pre}}$ from taking all encoded activity vectors except for $\mathbf{a}_u(t)$ as an input. Formally, the predicted activity is formulated as:

$$\mathbf{a}'_u(t) = \text{softmax}\left(\mathbf{W}_p \cdot \mathbf{h}_{\text{pre}} + \mathbf{b}_p\right), \quad (6)$$

where $\mathbf{W}_p \in \mathbb{R}^{k \times (n+7)}$ is the weight matrix for the final hidden state $\mathbf{h}_{\text{pre}}$, $k = 48$ is the hidden dimension, and $\mathbf{b}_p$ is the bias vector. We minimize the mean square error $\mathcal{L}_{\text{pre}}$ between the predicted activity $\mathbf{a}'_u(t)$ and the target activity $\mathbf{a}_u(t)$. The weights of the pre-trained model are then transferred to the learning outcome prediction module for training initialization. The dataset used in the pre-training process is aligned to that for training.

## 3.3 Meta-Learning Local Model Personalization

In order to facilitate personalized predictions based on student subgroup information, our methodology maintains a separate local model for each subgroup. The tasks handled locally are: (i) providing a subgroup-personalized model for global aggregations and (ii) adapting the global model based on the local subgroup dataset. We will investigate the impact of models constructed based on the different demographic variables $I \in \{G, C, Y\}$ from Section 2.4. In the following, we use $\mathcal{X}$ to denote an arbitrary set of subgroups for any particular variable. Further, we define the global prediction model as $\Theta_g$ and the local model for subgroup $x \in \mathcal{X}$ as $\Theta_x$.

We aim to train a global model that is *easily adaptable* to each local student subgroup. To this end, we employ a meta-learning based personalized FL framework, as in PerFed [16], seek the global model that solves the following optimization problem:

$$\min_{\Theta} \sum_{x \in \mathcal{X}} F_x(\Theta) \triangleq f_x\underbrace{\left(\Theta - \nabla f_x(\Theta)\right)}_{(a)}, \quad (7)$$

where $F_x(\cdot)$ is the meta-loss function of student subgroup $x$, and $f_x(\cdot)$ is the original loss function, which is the sum of prediction loss (i.e., BCE from Section 3.1) over the dataset of group $x$.

The main difference between our loss function in (7) and the loss functions used in existing student modeling work [6, 32] is that our loss function aims to minimize the loss of the *adapted* versions of the global model. This adaptation is based on one gradient descent step, i.e., term (a) in (7), which is taken over the local dataset of subgroup $x$. That is, our method exploits the *commonality* of data

---

**Algorithm 1** Attention-based Personalized Federated Learning

1: **Global Execution**:
2: Initialize global model $\Theta_g^{(0)}$
3: **for** each global aggregation period $k = 1, 2, ..., K$ **do**
4: 　　**for** each student subgroup $x \in \mathcal{X}$ **in parallel do**
5: 　　　　$\Theta_x^{(k,E)} \leftarrow$ **LocalAdaptation** ( $\Theta_g^{(k)}$ )
6: 　　**for** each subgroup $x \in \mathcal{X}$ **do**
7: 　　　　Compute attention weight $\alpha_x^{(k)}$ using (10).
8: 　　Obtain the global model $\Theta_g^{(k+1)}$ according to (11).

1: **LocalAdaptation** ( $\Theta_g^{(k)}$ ):
2: Initialize the local model $\Theta_x^{(k,0)} = \Theta_g^{(k)}$
3: **for** Each local iteration $e = 1, \cdots, E$ **do**
4: 　　Obtain $\Theta_x^{(k,e)}$ using the meta-update rule (8).
5: Return parameters $\Theta_x^{(k,E)}$

---

across subgroups to train an adaptable global model, which can be easily tailored to each individual subgroup.

To solve (7), we derive meta-gradient based local update steps. In particular, training proceeds through a sequence of training rounds $k \in \{1, \cdots, K\}$, with each round consisting of multiple local training iterations $e \in \{1, \cdots, E\}$. In each iteration $e$, subgroup $x \in \mathcal{X}$ updates its local model $\Theta_x^{(k,e)}$ using only the data of students from this subgroup. These local models are synchronized through a global aggregation step at the end of each round. Specifically, in each round $k$, the local model is initialized as $\Theta_x^{(k,0)} = \Theta_g^{(k)}$, $\forall x$, where $\Theta_g^{(k)}$ is the global aggregation at the end of $k - 1$. Then, each subgroup $x$ conducts its meta-gradient updates according to

$$\Theta_x^{(k,e)} = \Theta_x^{(k,e-1)} - \eta \nabla F_x\left(\Theta_x^{(k,e-1)}\right), \ e = 1, \cdots, E, \quad (8)$$

where $\eta$ is the step size, and based on (7) the meta gradient $\nabla F_x$ is

$$\nabla F_x(\Theta) = \left(\mathbf{I} - \nabla^2 f_x(\Theta)\right) \nabla f_x\left(\Theta - \nabla f_x(\Theta)\right), \ \forall \Theta, \quad (9)$$

where $\nabla^2$ is the Hessian operator. The second-order Hessian term in (9) can be well-approximated by first-order methods without incurring significant performance degradation, leading to efficient computations [16]. After these local updates, the global aggregation $\Theta_g^{(k+1)}$ is obtained using a method that we detail next.

## 3.4 Attention-Based Global Model Aggregation

The global modeling stage in our methodology is responsible for two tasks: (i) aggregating the local models and (ii) synchronizing the subgroup models with the resulting parameter vector for subsequent adaptation. Instead of using the standard averaging-based aggregation method from FedAvg [28], we employ a recently developed attention-based aggregation method, FedAtt [20]. FedAtt introduces an attention mechanism for weighting local neural network models according to how far the parameters in each of their layers have deviated from the previous global model.

Formally, at each aggregation step, the global model receives two pieces of information: (i) all the local models $\Theta_x^{(k,E)}$, and (ii) the attention weight for each local model. The attention weight

Yun-Wei Chu[†], Seyyedali Hosseinalipour[†], Elizabeth Tenorio[†], Laura Cruz[†], Kerrie Douglas[†], Andrew Lan[°], Christopher Brinton[†]

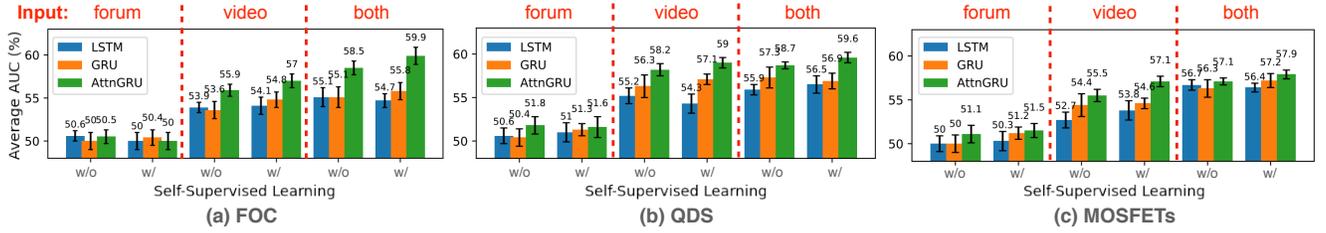

Figure 3: AUC scores of recurrent neural networks (LSTM, GRU, AttnGRU) with and without self-supervised behavioral pre-training according to different input modalities. AttnGRU performs the best among all the model structures, and self-supervised behavior learning equips models with better initializations yielding better performance in most cases. Incorporating forum activities with video activities improves the accuracy, showing that both activities are related to student knowledge acquisition.

$\alpha_x^{(k)}$ for local model $x$ is computed across its set of layers $\mathcal{L}$:

$$\alpha_x^{(k)} = \sum_{\ell \in \mathcal{L}} \text{softmax}\left(\left\|\Theta_g^{(k)}(\ell) - \Theta_x^{(k,E)}(\ell)\right\|\right), \quad (10)$$

where $\ell \in \mathcal{L}$ denotes a particular model layer, and $\Theta_g^{(k)}(\ell)$ and $\Theta_x^{(k,E)}(\ell)$ represent the parameter vector of the $\ell$-th layer in the global and local model, respectively, at the instance of global aggregation. The aggregated global model is then obtained as:

$$\Theta_g^{(k+1)} = \Theta_g^{(k)} - \epsilon \sum_{x \in \mathcal{X}} \alpha_x^{(k)} \left(\Theta_g^{(k)} - \Theta_x^{(k,E)}\right), \quad (11)$$

where $\epsilon$ is a tunable step size. The full attention-based personalized FL procedure is summarized in Algorithm 1 and Figure 2(c).

## 4 EXPERIMENTAL EVALUATION

We now conduct experiments on our three online course datasets from Section 2 to evaluate our personalization methodology.

### 4.1 Dataset Partitioning

Recall the subgroup datasets defined in Section 2.4. Instead of constructing personalizing models for only the students who provided the specific information $I \in \{G, C, Y\}$, in dataset $\Omega^I$, we also consider those who did not, in dataset $\Omega^{\neg I}$, as another student subgroup. For each subgroup under consideration, we first split students into train-test sets at a ratio of 4:1, that is, $\Omega^{I,x} = \{\Omega_{\text{train}}^{I,x}, \Omega_{\text{test}}^{I,x}\}$. Therefore, the overall training set is the union of the training set for each subgroup: $\Omega_{\text{train}}^I = \bigcup_{x \in \mathcal{X}} \Omega_{\text{train}}^{I,x}$; likewise, $\Omega_{\text{test}}^I = \bigcup_{x \in \mathcal{X}} \Omega_{\text{test}}^{I,x}$. In some cases, we will conduct our experiments on $\Omega^I = \{\Omega_{\text{train}}^I, \Omega_{\text{test}}^I\}$ only, representing the data of students who have provided their demographic information $I$. In other cases, we consider both $\Omega^I$ and $\Omega^{\neg I}$, where $\Omega^{\neg I} = \{\Omega_{\text{train}}^{\neg I}, \Omega_{\text{test}}^{\neg I}\}$.

### 4.2 Model Baselines and Variations

*4.2.1 Standard Recurrent Neural Networks:* While our main prediction model from Section 3.1 is an attention-based GRU model (AttnGRU), we also experiment with simpler recurrent models, LSTM and GRU. By taking the student activity vectors $\mathbf{a}_u(t)$ as the model input, LSTM and GRU then generate a final hidden state representing each student and predict the learning outcomes.

*4.2.2 Local Modeling:* We consider an attention-based GRU model, **AttnGRU-L**, which evaluates the efficacy of training separate local models for each student subgroup. More specifically, AttnGRU-L is trained on $\Omega_{\text{train}}^{I,x}$ and evaluated on $\Omega_{\text{test}}^{I,x}$ for each subgroup $x$.

*4.2.3 Global Modeling:* We implement three global models for comparison. One of them (G1) is a centralized model without FL, while the others (G2)&(G3) are Global FL models. All of them implement our proposed self-supervised behavioral pre-training process.

**(G1) AttnGRU-G**: This is a centralized global AttnGRU-based model that is trained on $\Omega_{\text{train}}^I$ and evaluated on $\Omega_{\text{test}}^I = \bigcup_{x \in \mathcal{X}} \Omega_{\text{test}}^{I,x}$.

**(G2) FedAvg [28]**: In FedAvg, for training round $k$, we train local models $\Theta_x^{(k,e)}$ on $\Omega_{\text{train}}^{I,x}$ without meta-learning. After $E$ iterations, FedAvg weighs each local model based on the number of students in each subgroup to conduct a standard global aggregation: $\Theta_g^{(k+1)} = \sum_{x \in \mathcal{X}} \frac{N_x}{N} \Theta_x^{(k,E)}$, where $N_x$ is the number of students within subgroup $x$, and $N = \sum_{x \in \mathcal{X}} N_x$. After $K$ global aggregations, we then use $\Theta_g^{(K)}$ to evaluate FedAvg on $\Omega_{\text{test}}^{I,x}$.

**(G3) FedAttn [20]**: After $E$ local iterations, FedAttn aggregates local models based on the attention mechanism with attention weights introduced in (10). After $K$ global aggregations, the aggregated global model $\Theta_g^{(K)}$ defined in (11) is evaluated on $\Omega_{\text{test}}^{I,x}$.

*4.2.4 Personalized Modeling:* We consider two baselines (P1)&(P2) with local model personalization in FL:

**(P1) FedIRT [40]**: As discussed in Section 1, [40] introduces federated deep knowledge tracing (FDKT) to coordinate between local and global models. They use classical test theory and item response theory (IRT) [34] to calculate the score representing the "data quality" of local subgroups, and use the scores to weight each subgroup when aggregating models.

Though we are not considering their specific knowledge tracing task, we implement their model update and aggregation method by calculating the IRT confidence $\alpha_x^{\text{IRT}}$ for each student subgroup $x$, since IRT produces the best results in their paper. Local model updates are based on hierarchical interpolation, where the initial model is defined as: $\Theta_x^{(k,0)} = \lambda^{(k)} \Theta_x^{(k-1,E)} + (1 - \lambda^{(k)}) \Theta_g^{(k)}$, $\lambda^{(k)} = \left(\Theta_x^{(k-1,E)} \cdot \Theta_g^{(k)}\right) / \left(\left\|\Theta_x^{(k-1,E)}\right\| \times \left\|\Theta_g^{(k)}\right\|\right)$, and the final local model $\Theta_x^{(k,E)}$ is obtained after $E$ local epochs of conventional gradient descent. The aggregated global model is then computed as $\Theta_g^{(k+1)} = \sum_{x \in \mathcal{X}} \alpha_x^{\text{IRT}} \Theta_x^{(k,E)}$. The IRT confidence in our setting is calculated using the students' answers to the end-of-video quizzes. For fair comparison with our personalization method, we use AttnGRU and apply our self-supervised pre-training.

**(P2) Ours¬AttnAgg**: Finally, to compare different aggregation rules that can be used in meta-learning, we also implement a baseline that replaces our attention-based aggregation method with the averaging-based method of [16]. Denoted Ours¬AttnAgg, this baseline also incorporates our self-supervised behavioral pre-training. The substantial performance improvements obtained by Ours¬AttnAgg over the other baselines will validate the benefit of meta-learning methods in model personalization.

| Dataset | | FOC | | QDS | | MOSFETs | |
|---|---|---|---|---|---|---|---|
| type | model | male | female | male | female | male | female |
| Local | AttnGRU-L | 54.6% (1.3%) | 60.0% (1.7%) | 62.5% (1.0%) | 61.2% (1.2%) | 60.3% (0.9%) | 55.5% (0.6%) |
| Global | AttnGRU-G | 59.7% (0.9%) | 62.7% (0.7%) | 60.3% (1.3%) | 59.4% (1.0%) | 59.2% (0.8%) | 56.4% (1.0%) |
|  | FedAvg | 56.3% (1.0%) | 61.9% (1.0%) | 57.7% (0.7%) | 60.3% (0.9%) | 56.3% (1.2%) | 57.5% (1.0%) |
|  | FedAttn | 57.2% (0.6%) | 62.2% (0.9%) | 59.2% (0.7%) | 59.6% (1.2%) | 56.9% (1.0%) | 56.2% (0.9%) |
| Personalized | FedIRT | 70.2% (1.3%) | 65.8% (0.9%) | 67.9% (1.5%) | 61.7% (1.3%) | 59.6% (1.2%) | 61.4% (1.0%) |
|  | Ours¬AttnAgg | 69.3%(0.6%) | **70.4%(0.5%)** | 71.4% (1.2%) | 68.0% (0.9%) | 64.9% (1.0%) | 63.8% (0.8%) |
|  | Ours | **73.5% (0.4%)** | **75.3% (1.3%)** | **79.5% (0.7%)** | **70.0% (0.9%)** | **66.1% (0.7%)** | **65.3% (0.9%)** |

**Table 3: Prediction performance obtained by different models on each gender subgroup. Overall, we see that our meta learning-based personalization methods obtain substantial improvements over the baselines, particularly for the FOC and QDS datasets.**

### 4.3 Implementation and Evaluation Metric

We perform 5-fold cross-validation on each dataset and randomly select 20% of the students in the training set for validation. For all models that use the attention-based GRU, the hidden state dimension is 48. 50 training epochs are used for non-FL approaches (LSTM, GRU, AttnGRU-L, and AttnGRU-G), and for FL methods (FedAvg, FedAttn, FedIRT, PerFed, and our method), the global aggregations ($K$) and local iterations ($E$) are set to be 10 and 5, respectively. For all models, the unfixed hyper-parameters are determined by a grid search over the search space: (i) Learning Rate: {1e-2, 5e-2, **1e-3**, 5e-3, 1e-4, 1e-5}, (ii) Learning Rate Decay: {5e-2, **1e-3**, 5e-3, 1e-4}, (iii) Droupout Rate: {0.3, 0.4, **0.5**, 0.6, 0.7}, (iv) Optimizer: {**Adam**, SGD}, and (v) Batch Size: {4, **8**, 16, 32}. Our method is trained using the parameters highlighted in boldface. We test all combinations of parameters for each baseline model and use the one achieving the highest accuracy in Section 4.4.

To evaluate the prediction performance, we use the Area Under the ROC Curve (AUC) metric. The ROC curve shows the true positive rate versus false positive rate in a binary classification task, and AUC assesses the tradeoff between these rates. Random guessing will result in an AUC of 0.5, while a perfect model has an AUC of 1. We execute each experiment five times with random initialization for each method and report the average and the standard deviation of AUC values. We evaluate on the test set using the model at the training epoch with the highest validation AUC value. For our method, according to (7), we measure the AUC of the adapted models obtained after performing one local epoch on the global model at each student subgroup.

### 4.4 Results and Discussion

*4.4.1 Model Structures and Input Features Analysis.* First, in Figure 3, we compare the performance of models using different recurrent neural networks (LSTM, GRU, and AttnGRU), models with and without self-supervised behavioral pre-training, and models using different sources of activity data. All models are trained on $\{\Omega^I_{\text{train}}, \Omega^{\neg I}_{\text{train}}\}$ and tested on $\{\Omega^I_{\text{test}}, \Omega^{\neg I}_{\text{test}}\}$ without grouping students. Considering only forum-participation activity ($\mathbf{a}^f$), all models tend to overfit due to the small amount of data, yielding approximately 50% AUC and high standard deviation for all model structures. Using only video-watching activity ($[\mathbf{v}^{id}; \mathbf{a}^v]$), all models consistently outperform those that use $\mathbf{a}^f$ only, which shows that video activities contain more information on knowledge acquisition than forum activities in our datasets. For AttnGRU without self-supervised behavioral pre-training, the average AUC improves by 5.5% when replacing forum activities with video activities. However, *incorporating both forum activity and video activity results in the best performance in most cases,* which shows that forum activities also contain useful information about students' learning processes and outcomes when incorporated properly. The average AUC improves 1.6% from using $[\mathbf{v}^{id}; \mathbf{a}^v]$ as compared to using $[\mathbf{a}^f; \mathbf{v}^{id}; \mathbf{a}^v]$ for AttnGRU without self-supervised behavioral pre-training.

The consistently better performance of AttnGRU over LSTM and GRU indicates that the attention mechanism can effectively incorporate activities in the distant past, especially for students with a large number of activities. Using both video and forum activities ($[\mathbf{a}^f; \mathbf{v}^{id}; \mathbf{a}^v]$) as input and without performing self-supervised pre-training, AttnGRU outperforms GRU and LSTM by an average AUC of 2.5% and 3.3%. Furthermore, *initializing the prediction model with self-supervised behavioral pre-training improves the performance in most cases,* revealing that pre-training gives us better representations of different student activity types. To be more specific, using both forum-participation and video-watching as input, the AUC in AttnGRU increases by 1.4% with the help of pre-training.

Apart from the listed RNN models, for completeness, we also conducted experiments using the Transformer model [38]. However, we did not observe significant improvement from replacing AttnGRU with Transformer architecture, which we attribute to the limited amount of student samples available in our setting. Recent work has also shown the promise of non Transformer-based models in different applications [35, 39]. Henceforth, we use our best model setting, i.e., AttnGRU with self-supervised pre-training that uses both video-watching and forum-participation activity as input.

*4.4.2 Predictive Quality on Student Subgroups.* In Tables 3, 4, and 5, we compare our method to the baselines in terms of the predictive quality achieved on each student group. We first show results on the set of students who provided specific demographic information $I$; we train models on $\Omega^I_{\text{train}}$ and test them on each subgroup of $\Omega^I_{\text{test}}$. Specifically, we divide the dataset into groups by gender, continent, and age information in Tables 3, 4, and 5, respectively. We also present the results of incorporating students who did not disclose their personal information (i.e., the models trained on $\{\Omega^I_{\text{train}}, \Omega^{\neg I}_{\text{train}}\}$ and tested on each subgroup of $\{\Omega^I_{\text{test}}, \Omega^{\neg I}_{\text{test}}\}$) in Tables 6, 8, and 7; the results are qualitatively similar.

In Tables 3, 4, and 5, we can see that AttnGRU-L provides only up to 60% AUC on each subgroup according to different demographic groupings, confirming our hypothesis that small subgroups do not Yun-Wei Chu[†], Seyyedali Hosseinalipour[†], Elizabeth Tenorio[†], Laura Cruz[†], Kerrie Douglas[†], Andrew Lan[◦], Christopher Brinton[†] have enough data for the model to be highly accurate. AttnGRU-L is thus not practical since local models cannot benefit from the *commonality* of information contained in other subgroups. Also, AttnGRU-G outperforms both global federated models (FedAvg and FedAttn) in most cases, suggesting that the centralized model delivers a better overall picture of the data than decentralized models when personalization is not employed. However, we also find that AttnGRU-G often performs better on subgroups with more students, which indicates that the model will ignore minorities and thus may have ethical concerns. For example, the performance of AttnGRU-G on the male subgroup is better that on the female subgroup for QDS and MOSFETs (see Table 3); moreover, the model performs better on the Asian and North American subgroups than other subgroups on FOC and QDS (see Table 4).

| Dataset | FOC | | | | |
|---|---|---|---|---|---|
| model | AS | AF | EU | NA | SA |
| AttnGRU-L | 57.3% (1.3%) | 60.1% (1.0%) | 61.5% (1.1%) | 53.9% (1.5%) | 60.7% (0.9%) |
| AttnGRU-G | 62.9% (1.1%) | 54.2% (0.7%) | 58.9% (1.5%) | 64.7% (1.4%) | 56.4% (1.0%) |
| FedAvg | 58.9% (0.6%) | 55.4% (0.7%) | 60.1% (1.2%) | 63.9% (1.1%) | 57.1% (1.1%) |
| FedAttn | 60.1% (1.1%) | 54.0% (1.0%) | 59.9% (0.7%) | 64.4% (0.5%) | 56.3% (0.9%) |
| FedIRT | 63.2% (0.8%) | 65.1% (1.0%) | 73.3% (0.6%) | 67.3% (0.8%) | 71.1% (0.9%) |
| Ours¬AttnAgg | 66.2% (1.0%) | 68.0% (1.1%) | 77.9% (0.8%) | 68.4% (1.1%) | 79.3% (0.9%) |
| Ours | 66.5% (0.8%) | 67.2% (1.0%) | 79.2% (1.1%) | 68.9% (0.9%) | 81.1% (1.3%) |
| Dataset | QDS | | | | |
| model | AS | AF | EU | NA | SA |
| AttnGRU-L | 61.2% (0.6%) | 64.7% (1.0%) | 57.3% (1.6%) | 59.9% (1.4%) | 62.8% (1.0%) |
| AttnGRU-G | 68.7% (1.0%) | 57.6% (0.9%) | 52.5% (1.1%) | 60.9% (1.4%) | 53.3% (1.1%) |
| FedAvg | 67.7% (1.2%) | 56.8% (1.5%) | 54.3% (0.8%) | 61.2% (1.0%) | 53.1% (1.2%) |
| FedAttn | 68.3% (1.0%) | 57.4% (0.8%) | 54.0% (0.8%) | 62.0% (1.4%) | 52.9% (1.1%) |
| FedIRT | 77.9% (0.8%) | 71.9% (1.1%) | 68.2% (0.9%) | 70.9% (1.2%) | 68.2% (1.5%) |
| Ours¬AttnAgg | 79.2% (1.3%) | 76.4% (1.5%) | 72.3% (1.3%) | 73.8% (1.6%) | 75.5% (1.0%) |
| Ours | 80.7% (1.0%) | 77.0% (1.1%) | 72.0% (0.6%) | 75.4% (0.9%) | 77.4% (1.0%) |
| Dataset | MOSFETs | | | | |
| model | AS | AF | EU | NA | SA |
| AttnGRU-L | 58.6% (1.1%) | 55.4% (0.9%) | 60.3% (0.7%) | 60.6% (1.0%) | 59.3% (0.9%) |
| AttnGRU-G | 61.4% (1.0%) | 52.3% (0.9%) | 55.7% (1.3%) | 56.4% (1.0%) | 59.1% (0.9%) |
| FedAvg | 60.5% (1.6%) | 55.8% (0.8%) | 58.1% (0.8%) | 55.3% (1.0%) | 58.3% (0.9%) |
| FedAttn | 60.8% (0.9%) | 54.9% (1.1%) | 58.9% (0.7%) | 54.7% (1.3%) | 58.7% (0.9%) |
| FedIRT | 63.1% (1.0%) | 65.3% (1.3%) | 63.7% (0.9%) | 61.2% (0.8%) | 58.4% (0.6%) |
| Ours¬AttnAgg | 62.8% (1.0%) | 69.4% (1.1%) | 62.8% (1.2%) | 64.9% (0.9%) | 60.5% (1.1%) |
| Ours | 63.9% (1.5%) | 69.9% (1.1%) | 64.3% (0.8%) | 65.4% (0.5%) | 60.1% (0.4%) |

**Table 4: Prediction performance obtained with different models on each student subgroup grouped by continent. AS, AF, EU, NA, and SA represent Asian, African, European, North American, and South American student subgroups. Our findings are consistent with Table 3, except for this subgroup, AttnAgg only provides marginal additional improvement.**

Across Tables 3, 4, and 5, we see that *most local- and global-based models have at least 10% lower AUC than the personalized models on the FOC and QDS datasets.* The performance of the personalized models is possibly limited in the MOSFETs dataset by its overall smaller size (see Table 1). The improvements obtained by the personalized models show the significance of adapting the global models based on local student subgroup data, which cannot be matched even by a sophisticated global aggregation technique such as FedAttn. Importantly, *our method and Ours¬AttnAgg consistently outperform FedIRT in most cases by at least 5%,* which suggests the benefits offered by our meta learning-based personalization approach. This improvement validates our hypothesis that weighting student subgroups based on their "data quality" as in FedIRT may lose important information on subgroup heterogeneity, which is captured by our redefinition of the global model in (7). Finally, compared to Ours¬AttnAgg, our method performs better on most student subgroups, albeit marginally for the country demographic. This confirms our intuition on the advantages of layer-wise model aggregations with an attention mechanism.

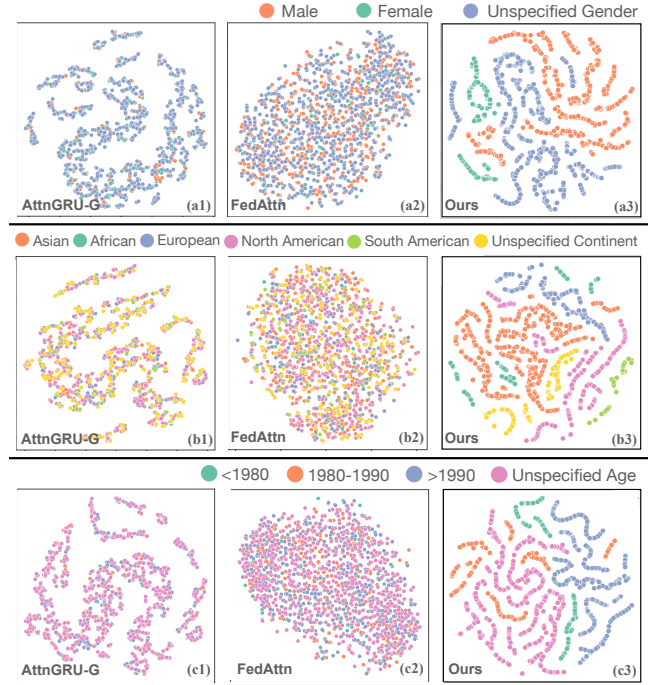

**Figure 4: Student representations learned by the centralized method (AttnGRU-G), attention-based federated learning (FedAttn), and our personalized method according to different demographic variables: (a1-3) gender, (b1-3) continent, and (c1-3) age. The organized and even clustered representation vectors learned by our method are consistent with its design to learn unique representations for each subgroup.**

Comparing Tables 6, 8, and 7 (incorporating students who did not disclose their personal information) with Tables 3, 4, and 5, we find a similar trends for the results. For example, centralized model (AttnGRU-G) performs better than federated global models (FedAvg and FedAttn), and personalized models outperform global- and local-based models. Our method consistently outperforms baselines on each subgroup, showing that personalized prediction with a better aggregation method can lead to a better prediction accuracy.

### 4.5 Embedding Visualization

In Figure 4, we use t-SNE to visualize the learned students' activity representations in a 2D space to qualitatively assess the intepretability of our framework. We plot the student embeddings for our method, FedAttn, and AttnGRU-G, according to different demographic groupings in the QDS dataset. All models are trained on $\{\Omega_{\text{train}}^{I}, \Omega_{\text{train}}^{\neg I}\}$, and the student embeddings are the combined hidden state $\widetilde{\mathbf{h}}_u$ after the attention module that is used to predict their learning outcomes. The colors represent the corresponding student demographic groups. We see that the student embeddings learned by our method are organized and sometimes well-clustered, reflecting specific demographic groups. Compared to AttnGRU-G and FedAttn, this is correlated with our method's better inference ability in predicting student performance. These visualizations confirm the ability of our method to adapt a global model to each student subgroup, helping us learn more expressive and predictive representations of student behavior, since we exploit available information unique to each subgroup that would be otherwise discarded.

| Dataset | | FOC | | | QDS | | | MOSFETs | | |
|---|---|---|---|---|---|---|---|---|---|---|
| type | model | ~80 | 80~90 | 90~ | ~80 | 80~90 | 90~ | ~80 | 80~90 | 90~ |
| Local | AttnGRU-L | 58.2% (1.3%) | 60.0% (1.7%) | 61.9% (1.4%) | 61.7% (1.0%) | 58.8% (1.0%) | 60.1% (0.9%) | 55.9% (1.2%) | 59.3% (1.1%) | 56.1% (0.9%) |
| Global | AttnGRU-G | 61.3% (1.0%) | 55.2% (1.2%) | 59.9% (1.1%) | 58.7% (0.7%) | 61.0% (0.4%) | 57.2% (0.6%) | 56.4% (1.1%) | 58.9% (0.9%) | 57.0% (0.9%) |
| | FedAvg | 62.2% (0.7%) | 54.3% (0.6%) | 58.6% (0.9%) | 61.9% (0.9%) | 60.3% (0.9%) | 58.6% (1.1%) | 55.3% (1.3%) | 58.6% (1.0%) | 55.2% (1.0%) |
| | FedAttn | 63.9% (1.1%) | 54.8% (0.9%) | 58.0% (0.8%) | 61.2% (0.9%) | 61.9% (1.0%) | 59.4% (1.0%) | 56.1% (1.0%) | 58.1% (0.9%) | 56.8% (1.2%) |
| Personalized | FedIRT | 69.7% (0.8%) | 79.9% (1.3%) | 71.3% (1.1%) | 65.4% (0.8%) | 73.8% (1.2%) | 68.9% (0.5%) | 60.9% (0.7%) | 62.3% (0.8%) | 65.4% (1.1%) |
| | Ours¬AttnAgg | 74.9% (1.2%) | 83.2% (2.3%) | 73.3% (1.6%) | 67.9% (1.0%) | 76.8% (1.2%) | 73.9% (1.4%) | 59.2% (1.1%) | 61.6% (1.2%) | 68.2% (1.5%) |
| | Ours | 75.2% (1.1%) | 83.0% (1.9%) | 75.9% (1.5%) | 67.2% (1.6%) | 78.9% (1.6%) | 75.1% (1.1%) | 61.3% (0.9%) | 62.5% (1.0%) | 68.2% (1.2%) |

Table 5: The prediction performance obtained with different models on each student subgroup grouped by age. ~80, 80 ~ 90, 90 ~ represent year of birth prior to 1980, between 1980 and 1990, and after 1990, respectively.

| Dataset | | FOC | | | QDS | | | MOSFETs | | |
|---|---|---|---|---|---|---|---|---|---|---|
| type | model | male | female | unspecified | male | female | unspecified | male | female | unspecified |
| Local | AttnGRU-L | 54.6% (1.0%) | 60.0% (1.1%) | 56.2% (0.9%) | 62.5% (1.3%) | 61.2% (1.1%) | 61.7% (1.3%) | 58.4% (0.8%) | 59.7% (0.9%) | 60.4% (1.1%) |
| Global | AttnGRU-G | 62.2% (1.3%) | 55.9% (0.7%) | 57.1% (1.5%) | 59.2% (0.9%) | 63.4% (1.3%) | 58.8% (1.0%) | 59.9% (1.0%) | 60.3% (1.2%) | 61.9% (1.1%) |
| | FedAvg | 58.3% (0.9%) | 62.1% (1.2%) | 56.5% (1.0%) | 58.8% (1.5%) | 59.3% (1.0%) | 59.9% (1.1%) | 62.2% (1.0%) | 58.3% (0.9%) | 59.2% (1.1%) |
| | FedAttn | 59.4% (1.0%) | 63.5% (1.2%) | 57.3% (1.1%) | 60.1% (1.0%) | 62.0% (0.7%) | 58.6% (0.9%) | 61.8% (0.6%) | 58.9% (1.0%) | 59.9% (1.0%) |
| Personalized | FedIRT | 62.0% (1.6%) | 61.0% (1.3%) | 55.4% (0.9%) | 60.9% (1.1%) | 64.9% (1.2%) | 59.4% (1.0%) | 65.2% (1.3%) | 69.3% (1.1%) | 66.4% (1.5%) |
| | Ours¬AttnAgg | 66.3% (1.0%) | 68.4% (1.2%) | 66.3% (1.2%) | 73.9% (1.6%) | 64.3% (0.9%) | 65.8% (1.5%) | 70.9% (1.0%) | 71.5% (1.1%) | 65.8% (0.9%) |
| | Ours | 67.9% (0.8%) | 69.1% (1.0%) | 68.8% (1.5%) | 73.1% (1.1%) | 65.7% (1.0%) | 65.6% (0.9%) | 71.2% (1.1%) | 71.5% (1.0%) | 66.4% (1.2%) |

Table 6: The prediction performance obtained with different methods on each student subgroup grouped by gender. The "unspecified" subgroup represents the students who did not disclose their gender. Similar to Table 3, the results show that our method outperform baselines on most subgroups.

| Dataset | | FOC | | | | QDS | | | | MOSFETs | | | |
|---|---|---|---|---|---|---|---|---|---|---|---|---|---|
| type | model | ~80 | 80~90 | 90~ | unspecified | ~80 | 80~90 | 90~ | unspecified | ~80 | 80~90 | 90~ | unspecified |
| Local | AttnGRU-L | 58.2% (1.0%) | 60.0% (1.7%) | 61.9% (0.9%) | 55.3% (0.9%) | 61.7% (1.0%) | 58.8% (1.1%) | 60.1% (0.9%) | 55.4% (1.1%) | 62.9% (0.9%) | 55.3% (1.0%) | 55.9% (1.0%) | 59.2% (1.1%) |
| Global | AttnGRU-G | 63.5% (1.4%) | 59.7% (1.1%) | 62.6% (1.0%) | 61.9% (1.0%) | 59.6% (1.2%) | 60.1% (0.9%) | 61.2% (1.7%) | 58.4% (0.8%) | 61.3% (1.0%) | 62.9% (1.0%) | 58.3% (0.9%) | 59.8% (0.9%) |
| | FedAvg | 59.2% (1.0%) | 58.9% (1.1%) | 61.2% (0.9%) | 59.4% (1.1%) | 60.9% (0.7%) | 60.9% (0.7%) | 59.7% (1.3%) | 59.6% (1.0%) | 60.2% (0.8%) | 59.2% (0.9%) | 58.6% (1.2%) | 55.9% (0.8%) |
| | FedAttn | 60.6% (1.3%) | 58.3% (0.9%) | 61.5% (1.0%) | 58.8% (1.1%) | 59.1% (1.0%) | 62.3% (1.6%) | 60.1% (0.9%) | 61.3% (1.2%) | 60.2% (0.8%) | 61.3% (1.0%) | 56.0% (1.1%) | 56.5% (1.2%) |
| Personalized | FedIRT | 69.2% (1.7%) | 74.2% (2.0%) | 76.0% (2.7%) | 65.2% (1.5%) | 70.9% (1.6%) | 72.8% (1.8%) | 70.1% (0.9%) | 70.5% (1.1%) | 69.3% (1.3%) | 66.9% (1.0%) | 62.4% (1.2%) | 67.2% (1.1%) |
| | Ours¬AttnAgg | 70.8% (2.1%) | 75.2% (2.0%) | 75.0% (1.9%) | 64.2% (1.1%) | 70.9% (2.0%) | 73.1% (1.0%) | 74.0% (2.2%) | 72.5% (2.1%) | 73.2% (2.3%) | 70.2% (1.9%) | 68.4% (1.1%) | 71.5% (2.1%) |
| | Ours | 71.2% (1.8%) | 77.1% (2.0%) | 77.1% (1.5%) | 65.1% (1.6%) | 71.4% (1.4%) | 73.3% (2.0%) | 75.1% (1.7%) | 72.3% (1.0%) | 74.8% (1.7%) | 69.5% (1.2%) | 69.6% (0.9%) | 70.1% (1.6%) |

Table 7: The prediction performance obtained with different methods on each student subgroup grouped by age. The "unspecified" subgroup represents the students who did not disclose their year of birth.

In addition, the visuals show that student embeddings are at least moderately correlated with the specific student subgroup. For instance, in case of gender (Figure 4(a3)), males (orange dots) tend to appear on the right side of the figure, while females (turquoise dots) tend to be at the far left. The same phenomenon can be observed in other subplots of Figure 4. This result suggests that students from different subgroups exhibit some unique learning behaviors, consistent with prior work discussed in Section 1.2. Our meta learning-based personalization approach takes into account via tailoring global models to different student subgroups.

## 5 CONCLUSION

In this paper, we developed a personalized federated learning framework for improving student modeling on underrepresented groups. Our approach is based on meta learning, where local models corresponding to different student subgroups are adapted from a shared global crafted to incorporate the personalization step. We applied our framework to the problem of predicting learning outcomes from student activities, where subgroups are defined according to demographic variables. Our evaluation on three online course datasets showed that our methodology (sometimes significantly) outperforms baseline algorithms on maximizing predictive quality for *every* student subgroup. Moreover, well-organized student embeddings learned by our method are correlated with improved student modeling. Avenues for future work include (i) exploring more rigorous student subgroup definitions [22] and (ii) applying our framework to other types of student models.

| Dataset | FOC | | | | | |
|---|---|---|---|---|---|---|
| model | AS | AF | EU | NA | SA | unspecified |
| AttnGRU-L | 57.3% (1.2%) | 60.1% (0.9%) | 61.5% (1.6%) | 53.9% (1.5%) | 60.7% (1.1%) | 54.8% (1.0%) |
| AttnGRU-G | 59.4% (1.1%) | 53.1% (1.0%) | 58.3% (0.9%) | 61.3% (1.1%) | 61.2% (1.1%) | 58.7% (0.6%) |
| FedAvg | 58.1% (0.9%) | 60.7% (0.9%) | 54.6% (1.1%) | 60.9% (1.0%) | 60.1% (0.9%) | 55.2% (1.1%) |
| FedAttn | 58.9% (1.4%) | 59.9% (0.9%) | 56.3% (1.1%) | 59.0% (1.3%) | 60.8% (1.5%) | 56.9% (1.4%) |
| FedIRT | 62.3% (0.9%) | 70.2% (1.2%) | 69.9% (1.4%) | 65.8% (1.1%) | 63.3% (1.5%) | 57.0% (1.6%) |
| Ours¬AttnAgg | 61.9% (1.3%) | 73.1% (1.1%) | 71.2% (1.5%) | 66.7% (1.2%) | 65.2% (1.1%) | 57.9% (1.2%) |
| Ours | 62.3% (0.9%) | 73.9% (1.2%) | 73.6% (1.8%) | 66.4% (1.6%) | 66.0% (1.0%) | 58.4% (1.1%) |
| Dataset | QDS | | | | | |
| model | AS | AF | EU | NA | SA | unspecified |
| AttnGRU-L | 61.2% (1.2%) | 64.7% (1.0%) | 57.3% (0.9%) | 59.9% (1.0%) | 62.8% (1.2%) | 53.2% (0.7%) |
| AttnGRU-G | 60.5% (0.9%) | 62.3% (1.1%) | 55.6% (1.1%) | 59.0% (0.9%) | 61.0% (1.2%) | 57.9% (1.0%) |
| FedAvg | 63.1% (1.1%) | 60.9% (1.0%) | 58.7% (1.2%) | 59.2% (0.9%) | 59.3% (1.1%) | 54.9% (0.9%) |
| FedAttn | 64.3% (0.9%) | 59.2% (1.0%) | 60.0% (1.1%) | 58.9% (1.6%) | 60.2% (0.9%) | 56.6% (1.3%) |
| FedIRT | 68.9% (1.1%) | 71.3% (1.7%) | 72.3% (1.3%) | 70.9% (1.8%) | 67.2% (0.8%) | 55.4% (1.4%) |
| Ours¬AttnAgg | 73.8% (0.7%) | 80.4% (1.9%) | 74.9% (1.1%) | 72.3% (1.6%) | 66.1% (1.5%) | 57.1% (1.1%) |
| Ours | 75.5% (0.8%) | 78.3% (2.0%) | 75.5% (1.1%) | 72.8% (1.9%) | 68.2% (1.0%) | 57.9% (1.2%) |
| Dataset | MOSFETs | | | | | |
| model | AS | AF | EU | NA | SA | unspecified |
| AttnGRU-L | 59.4% (0.5%) | 55.3% (1.0%) | 57.2% (1.3%) | 60.8% (0.9%) | 55.3% (1.0%) | 58.6% (0.9%) |
| AttnGRU-G | 63.2% (1.1%) | 60.9% (1.0%) | 58.1% (1.2%) | 56.7% (0.9%) | 53.6% (1.1%) | 59.8% (1.0%) |
| FedAvg | 60.2% (1.2%) | 58.9% (1.0%) | 56.3% (1.1%) | 55.8% (1.3%) | 54.1% (1.3%) | 58.2% (1.6%) |
| FedAttn | 60.5% (0.8%) | 58.5% (1.1%) | 55.9% (1.6%) | 57.1% (0.7%) | 53.6% (1.3%) | 58.5% (1.1%) |
| FedIRT | 66.3% (1.0%) | 68.3% (1.1%) | 70.5% (1.5%) | 69.4% (1.1%) | 68.8% (0.9%) | 69.2 (1.2%) |
| Ours¬AttnAgg | 68.9% (1.0%) | 67.9% (1.1%) | 69.2% (1.5%) | 72.5% (1.9%) | 69.4% (1.1%) | 72.3% (1.6%) |
| Ours | 69.4% (1.8%) | 68.5% (1.1%) | 70.1% (1.9%) | 73.3% (2.0%) | 69.4% (1.5%) | 72.9% (1.3%) |

Table 8: The performance of different models on each student subgroup by continent demographic variable. The "unspecified" subgroup represents the students who did not disclose their continent information.

# REFERENCES

[1] Olasile Babatunde Adedoyin and Emrah Soykan. 2020. Covid-19 pandemic and online learning: the challenges and opportunities. *Interactive Learning Environments* (2020), 1–13.
[2] Stepfanie M. Aguillon, Gregor-Fausto Siegmund, Renee H. Petipas, Abby Grace Drake, Sehoya Cotner, and Cissy J. Ballen. 2020. Gender Differences in Student Participation in an Active-Learning Classroom. *CBE Life Sci. Edu.* 19 (2020).
[3] Julia Angwin and Hannes Grassegger. 2017. Facebook's secret censorship rules protect white men from hate speech but not black children. *PROPUBLICA* 28 (2017).
[4] Jonathan Bassen, Bharathan Balaji, Michael Schaarschmidt, Candace Thille, Jay Painter, Dawn Zimmaro, Alex Games, Ethan Fast, and John C Mitchell. 2020. Reinforcement learning for the adaptive scheduling of educational activities. In *CHI Conf. Human Factors Comput. Syst.* 1–12.
[5] Shai Ben-David, John Blitzer, K. Crammer, A. Kulesza, Fernando C Pereira, and Jennifer Wortman Vaughan. 2009. A theory of learning from different domains. *Machine Learning* 79 (2009), 151–175.
[6] Anthony F Botelho, Ryan S Baker, and Neil T Heffernan. 2017. Improving sensor-free affect detection using deep learning. In *Int. Conf. Artif. Intell. Edu.* 40–51.
[7] Christopher G. Brinton, Swapna Buccapatnam, M. Chiang, and H. Poor. 2016. Mining MOOC Clickstreams: Video-Watching Behavior vs. In-Video Quiz Performance. *IEEE Trans. Signal Process.* 64 (2016), 3677–3692.
[8] Christopher G. Brinton, Swapna Buccapatnam, Liang Zheng, Da Cao, Andrew S. Lan, Felix Ming Fai Wong, Sangtae Ha, Mung Chiang, and H. Vincent Poor. 2018. On the Efficiency of Online Social Learning Networks. *IEEE/ACM Trans. Netw.* 26 (2018), 2076–2089.
[9] Christopher G. Brinton and M. Chiang. 2015. MOOC performance prediction via clickstream data and social learning networks. *IEEE Int. Conf. Computer Comms.* (2015), 2299–2307.
[10] Joy Buolamwini and Timnit Gebru. 2018. Gender shades: Intersectional accuracy disparities in commercial gender classification. In *Conf. Fairness, Accountability and Transparency.* 77–91.
[11] Wai-Lun Chan and Dit-Yan Yeung. 2021. Clickstream Knowledge Tracing: Modeling How Students Answer Interactive Online Questions. In *Int. Learn. Analytics and Knowl. Conf.* 99–109.
[12] Shivangi Chopra, Hannah Gautreau, Abeer Khan, Melicaalsadat Mirsafian, and Lukasz Golab. 2018. Gender Differences in Undergraduate Engineering Applicants: A Text Mining Approach. In *EDM*.
[13] Yun-Wei Chu, Elizabeth Tenorio, Laura Cruz, Kerrie Anna Douglas, Andrew S. Lan, and Christopher G. Brinton. 2021. Click-Based Student Performance Prediction: A Clustering Guxided Meta-Learning Approach. *IEEE Int. Conf. Big Data* (2021), 1389–1398.
[14] Albert T Corbett and John R Anderson. 1994. Knowledge tracing: Modeling the acquisition of procedural knowledge. *User modeling and user-adapted interaction* 4, 4 (1994), 253–278.
[15] Corinna Cortes and M. Mohri. 2014. Domain adaptation and sample bias correction theory and algorithm for regression. *Theor. Comput. Sci.* 519 (2014), 103–126.
[16] Alireza Fallah, Aryan Mokhtari, and A. Ozdaglar. 2020. Personalized Federated Learning: A Meta-Learning Approach. *ArXiv* abs/2002.07948 (2020).
[17] Y Fan, LJ Shepherd, E Slavich, D Waters, M Stone, R Abel, and EL Johnston. 2019. Gender and cultural bias in student evaluations: Why representation matters. *PloS ONE* 14, 2 (2019), e0209749.
[18] Josh Gardner, Christopher Brooks, and Ryan Baker. 2019. Evaluating the fairness of predictive student models through slicing analysis. In *Int. Conf. Learn. Analyt. and Knowl.* 225–234.
[19] Yutao Huang, Lingyang Chu, Zirui Zhou, Lanjun Wang, Jiangchuan Liu, Jian Pei, and Yong Zhang. 2021. Personalized cross-silo federated learning on non-iid data. In *AAAI Conf. Artif. Intell.*, Vol. 35. 7865–7873.
[20] Shaoxiong Ji, Shirui Pan, Guodong Long, Xue Li, Jing Jiang, and Zi Huang. 2019. Learning Private Neural Language Modeling with Attentive Aggregation. *Int. Joint Conf. Neural Netw.* (2019), 1–8.
[21] Yihan Jiang, Jakub Konečný, Keith Rush, and S. Kannan. 2019. Improving Federated Learning Personalization via Model Agnostic Meta Learning. *ArXiv* abs/1909.12488 (2019).
[22] Michael Kearns, Seth Neel, Aaron Roth, and Zhiwei Steven Wu. 2018. Preventing fairness gerrymandering: Auditing and learning for subgroup fairness. In *Int. Conf. Machine Learn.* PMLR, 2564–2572.
[23] René F Kizilcec and Hansol Lee. 2020. Algorithmic fairness in education. *arXiv preprint arXiv:2007.05443* (2020).
[24] Jakub Konečný, H. B. McMahan, F. Yu, Peter Richtárik, A. T. Suresh, and D. Bacon. 2016. Federated Learning: Strategies for Improving Communication Efficiency. *ArXiv* abs/1610.05492 (2016).
[25] Preethi Lahoti, Krishna P Gummadi, and Gerhard Weikum. 2019. ifair: Learning individually fair data representations for algorithmic decision making. In *Int. Conf. Data Engrg.* IEEE, 1334–1345.
[26] Robert V Lindsey, Jeffery D Shroyer, Harold Pashler, and Michael C Mozer. 2014. Improving students' long-term knowledge retention through personalized review. *Psychological Science* 25, 3 (2014), 639–647.
[27] Jessica McBroom, Irena Koprinska, and Kalina Yacef. 2020. How Does Student Behaviour Change Approaching Dropout? A Study of Gender and School Year Differences. In *EDM*.
[28] H. B. McMahan, Eider Moore, D. Ramage, S. Hampson, and B. A. Y. Arcas. 2017. Communication-Efficient Learning of Deep Networks from Decentralized Data. In *Int. Conf. Artif. Intell. and Stats.*
[29] Tomas Mikolov, Kai Chen, G. Corrado, and J. Dean. 2013. Efficient Estimation of Word Representations in Vector Space. In *Int. Conf. Learn. Represent.*
[30] Connor Neill, Sehoya Cotner, M. Driessen, and Cissy J. Ballen. 2019. Structured learning environments are required to promote equitable participation. *Chemistry Education Research and Practice* (2019).
[31] Luc Paquette, Jaclyn Ocumpaugh, Ziyue Li, Alexandra Andres, and Ryan Baker. 2020. Who's Learning? Using Demographics in EDM Research. *Journal of Educational Data Mining* 12, 3 (2020), 1–30.
[32] C. Piech, J. Bassen, J. Huang, S. Ganguli, M. Sahami, L. Guibas, and J. Sohl-Dickstein. 2015. Deep knowledge tracing. In *Adv. Neural Inf. Process. Syst.* 505–513.
[33] Siddharth Reddy, Igor Labutov, Siddhartha Banerjee, and Thorsten Joachims. 2016. Unbounded human learning: Optimal scheduling for spaced repetition. In *ACM SIGKDD Int. Conf. Knowl. Disc. Data Mining.* 1815–1824.
[34] M. Tatsuoka, F. Lord, M. R. Novick, and A. Birnbaum. 1968. Statistical Theories of Mental Test Scores. *J. Amer. Statist. Assoc.* 66 (1968), 651.
[35] Yi Tay, Mostafa Dehghani, Jai Gupta, Dara Bahri, V. Aribandi, Zhen Qin, and Donald Metzler. 2021. Are Pre-trained Convolutions Better than Pre-trained Transformers? *ArXiv* abs/2105.03322 (2021).
[36] Wim J van der Linden and Ronald K Hambleton. 2013. *Handbook of modern item response theory.* Springer Science and Business Media.
[37] Kurt VanLehn. 1988. Student modeling. *Found. Intell. Tut. Syst.* 55 (1988), 78.
[38] Ashish Vaswani, Noam M. Shazeer, Niki Parmar, Jakob Uszkoreit, Llion Jones, Aidan N. Gomez, Lukasz Kaiser, and Illia Polosukhin. 2017. Attention is All you Need. *ArXiv* abs/1706.03762 (2017).
[39] Felix Wu, Angela Fan, Alexei Baevski, Yann Dauphin, and Michael Auli. 2019. Pay Less Attention with Lightweight and Dynamic Convolutions. *ArXiv* abs/1901.10430 (2019).
[40] Jinze Wu, Zhenya Huang, Qi Liu, Defu Lian, H. Wang, Enhong Chen, Haiping Ma, and Shijin Wang. 2021. Federated Deep Knowledge Tracing. *ACM Int. Conf. Web Search and Data Mining* (2021).
[41] Diyi Yang, Tanmay Sinha, David Adamson, and Carolyn Penstein Rosé. 2013. Turn on, tune in, drop out: Anticipating student dropouts in massive open online courses. In *Data-driven Educ. WRKSH Conf. Neural Inf. Process. Syst.*, Vol. 11. 14.
[42] Mengfan Yao, Siqian Zhao, Shaghayegh Sahebi, and Reza Feyzi Behnagh. 2021. Stimuli-sensitive Hawkes processes for personalized student procrastination modeling. In *The World Wide Web Conf.* 1562–1573.
[43] Sirui Yao and Bert Huang. 2017. Beyond parity: Fairness objectives for collaborative filtering. *arXiv preprint arXiv:1705.08804* (2017).
[44] Jiani Zhang, Xingjian Shi, Irwin King, and D. Yeung. 2017. Dynamic Key-Value Memory Networks for Knowledge Tracing. *The World Wide Web Conf.* (2017).